%% file: arxiv.tex
\documentclass{article}
\usepackage{spconf,amsmath,epsfig}
\usepackage{graphicx}
\usepackage{wrapfig}
\usepackage{subcaption}
\usepackage{float}
\usepackage[export]{adjustbox}
\usepackage{hyperref}
\usepackage{tikz}
\usepackage{standalone}
\usepackage{pgfplots}
\usetikzlibrary{calc}

\title{PREDICTING ELECTRICITY OUTAGES CAUSED BY CONVECTIVE STORMS}
\twoauthors
  {Roope Tervo, Joonas Karjalainen
  }
    {Observing and information systems centre\\
    Finnish Meteorological Institute \\
	 B.O. 503, 00101 Helsinki, Finland}
  {Alexander Jung 
  }
	{Dept of Computer Science\\
	Aalto University\\
	B.O. 11000, 00076 Aalto, Finland}

\newcommand\copyrighttext{
  \footnotesize \textcopyright {Copyright 2018 IEEE. Published in the 2018 IEEE Data Science Workshop (DSW 2018), scheduled for June 4-6, 2018 in Lausanne, Switzerland. Personal use of this material is permitted. However, permission to reprint/republish this material for advertising or promotional purposes or for creating new collective works for resale or redistribution to servers or lists, or to reuse any copyrighted component of this work in other works, must be obtained from the IEEE. Contact: Manager, Copyrights and Permissions / IEEE Service Center / 445 Hoes Lane / P.O. Box 1331 / Piscataway, NJ 08855-1331, USA. Telephone: + Intl. 908-562-3966. }}
\renewcommand\copyrightnotice{%
\begin{tikzpicture}[remember picture,overlay]
\node[anchor=south,yshift=10pt] at (current page.center) {\fbox{\parbox{\dimexpr\textwidth-\fboxsep-\fboxrule\relax}{\copyrighttext}}};
\end{tikzpicture}%
}

\begin{document}

\copyrightnotice

\maketitle

\begin{abstract}
We consider the problem of predicting power outages in an electrical power grid due to hazards produced by convective storms. These storms produce extreme weather phenomena such as intense wind, tornadoes and lightning over a small area. 
In this paper, we discuss the application of state-of-the-art machine learning techniques, such as random forest classifiers and deep neural networks, to predict the amount of damage caused by storms. We cast this application 
as a classification problem where the goal is to classify storm cells into a finite number of classes, each corresponding to a certain amount of expected damage. 
The classification method use as input features estimates for storm cell location and movement which has to be extracted from the raw data. 

A main challenge of this application is that the training data is heavily imbalanced as the occurrence of extreme weather events is rare. 
In order to address this issue, we applied SMOTE technique. 
\end{abstract}

\begin{keywords}
Power distribution, Weather Impact, Random Forest Classifier, Multilayer Perceptron, Neural Network
\end{keywords}
\section{Introduction}
\label{sec:intro}
A key problem faced by operators of power grids is the prediction of damages caused by extreme weather events. 
The hazards produced by intense winds, lightning and tornadoes have significant social impacts and cause remarkable liability for damages for distribution companies.
The main objective of this work is to apply machine learning techniques in order to obtain a real-time prediction of the short-term damage potential based on current weather information. 


As the weather-caused damages incur significant economic loss, a lot of effort has been put into studying efficient prediction methods.
The problem of storm cell identification and tracking has been studied thoroughly in \cite{rossi2015object}. We will use the methods developed in \cite{rossi2015object} as pre-processing of raw weather data.
The authors of \cite{doi:10.1175/BAMS-D-16-0123.1} provide an excellent overview of several successful projects. In general, random forest classifiers (RFC) \cite{breiman2001random} are very popular method for analysing weather impacts. The authors of \cite{kankanala2014adaboost} applied a variant of AdaBoost\textsuperscript{+} to predict outages caused by wind and lightning in overhead distribution systems. 

This paper is organised as follows: In Section \ref{sec:Problemsetup}, we formalise the problem of predicting outages as a classification problem using estimates of storm cell location and movements as input features. We then discuss the application of RFC and deep neural network classifiers in Section \ref{sec:Method}. Some illustrative numerical experiments based on historic data collected by the Finnish Meteorological Institute (FMI) are discussed in Section \ref{sec:Experiments}.


 \section{Problem Setup} 
 \label{sec:Problemsetup}
 

We identify convective storm cells by contouring weather radar reflectivity composite CAPPI (constant altitude plan position indicator) images with a solid 35 DBZ threshold. After contouring, we apply the DBSCAN method \cite{ester1996density} to cluster the contoured objects to an identifiable storm object which are stored to a PostGIS database. 

Within the DBSCAN method, a storm object is considered as a core point if the area sum of its nearby storm objects exceeds the given area threshold. The storm objects that are within the neighbourhood of a core point but do not fulfil the criteria mentioned above, are considered as outliers. Together with their outliers, connected core points form a cluster and storm objects that belong to neither of these classes are regarded as noise. In this study, the area limit was set to $20 {\rm km}^2$ and the neighbourhood radius was set to 
$2 {\rm km}^{2}$. After clustering, an object-oriented storm tracking algorithm using optical flow \cite{horn1981determining} is used to track and forecast paths of the storms. The storm cell tracking is done for a time horizon of 2 hours ahead with a time resolution of 5 minutes.

In order to recognise damage potential of convective storms and to predict future outages, we categorise storm cells into four classes based on how much damage they are expected to cause for the power grid. In more detail, the storm cells are assigned to a class based on how large share of transformers under the storm are without electricity. The damage may happen to any point in the grid, but outages are always reported at the transformer nodes of the power grid.

Thus, we formulate the outage prediction as a classification problem which aims at classifying storm cells into one of four different classes, described in table \ref{table:classes}.  The particular choice of this classes aims to provide a simple 'at glance' view which is convenient for the end user (power grid operator).

\newlength\e
\setlength\e{\dimexpr .15\columnwidth -2\tabcolsep}
\newlength\ee
\setlength\ee{\dimexpr .50\columnwidth -2\tabcolsep}
\begin{table}[h]
    \centering
    \caption{Class definitions of the storm cells.}
    \label{table:classes}
    \begin{tabular}[width=0.5\columnwidth]{ p{\e} p{\ee} } 
    \textbf{Class} & \textbf{Share of transformers } \\
    \hline
    0 & no damage \\ 
    \hline
    1 & 0 - 10 \% \\ 
    \hline
    2 & 10 - 50 \% \\ 
    \hline
    3 & 50 - 100 \% \\ 
    \end{tabular}
\end{table}

We used data collected by FMI during years 2012 to 2017 to train the classification methods. This data contains identified storm cell objects at a time resolution of five minutes. Note that while most of the time steps do not contain any storm cells, during ``active'' days, each time step may contain hundreds of cells. Fetching all storm cell objects covering the power grid at some point of their life cycle provided 886 020 training samples.

Each training sample represents one individual storm cell which is characterized by the list of features listed in Table \ref{table:input_data}. The data is very imbalanced as most of the convective cells are not powerful enough to cause harm for the power grid. We depict a histogram of the target classes contained in the training dataset in Figure \ref{fig:y_hist_syntetic}. In particular, we have 551 029 samples of class 0 (no harm), 4919 samples of class 1, 4286 samples belonged to class 2 and only 3337 belong to class 3 (most harmful).

\input{param_table.tex}

Many ground observations were missing as they were fetched from the nearest observation station to the storm center and capabilities of those weather stations varies a lot. Absent parameters were initialised to zero to ensure coherence of the data. Effects of this inadequate data is discussed more in section \ref{sec:Experiments}.

\begin{figure}[t]
\centering
\includegraphics[width=0.7\linewidth]{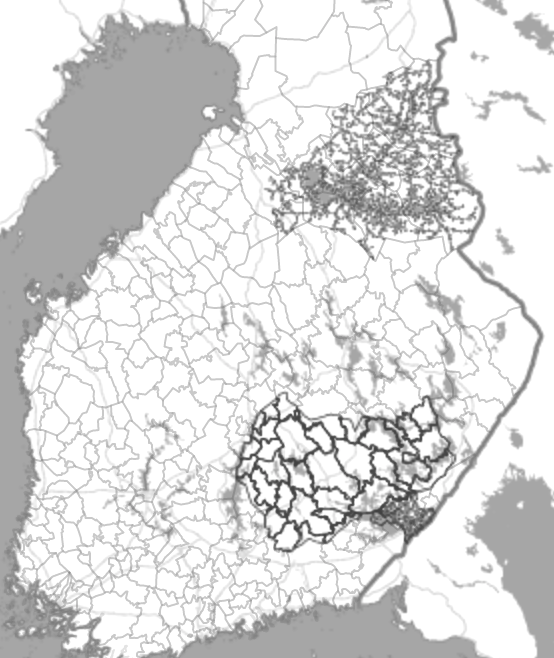} 
\caption{Spatial coverage of power grid information available in this project. Soft lines represents municipality districts and darker lines power grids    . Image is from Southern Finland.}
\label{fig:network_area}
\vspace{-10pt}
\end{figure}

Outage data and power grid description are fetched from two power distribution companies. Spatial coverage of the data is shown in image \ref{fig:network_area}. The data set contains in total 33 858 outages. 

\section{The Classification Method}\label{sec:Method}


We created two alternative methods for classification. First, we used RFC to get 'baseline' performance. The classifier was implemented without limitations in tree size and with equal class weights. For the training of the RFC, we used the Gini impurity as loss function, i.e.
\begin{equation}
G = -\sum_{i=1}^{n_c}(p_i(1-p_i))
\vspace*{-1mm}
\end{equation}
where $n_c$ is the number of classes and $p_i$ is the share of $i^{th}$ class in the tree.

\begin{figure}[ht]
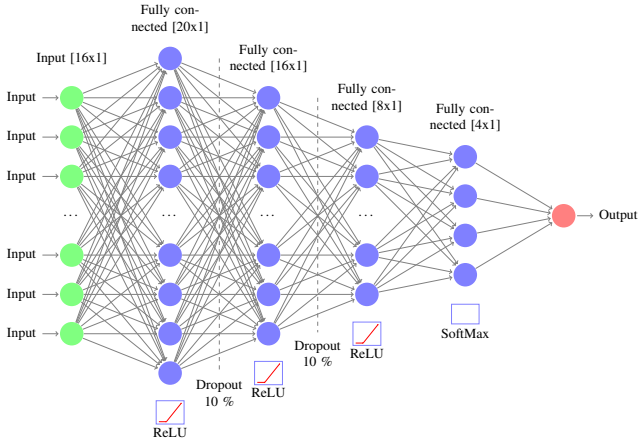

\centering
\includestandalone[width=\linewidth]{nn_struct}
\caption{Network structure for classification task}
\label{fig:model_class}
\end{figure}

As an alternative to RFC, we also implemented a classifier based on a multi layer perceptron (MLP) neural network \cite{Goodfellow-et-al-2016-mlp}. The network structure is described in Figure \ref{fig:model_class}. The first layer contains 20 nodes wide dense layer. In the following layers, the number of nodes is reduced to 16, 8, 4 and finally to 1 node. The first three dense layers use the rectified linear unit (Relu) as activation and dropout regularisation layers are included after first and second dense layers. In the final layer, we used the ``Softmax'' activation function in order to obtain the predicted class probabilities in the output layer.

In order to cope with the imbalanced training data we decided to use the cross entropy loss function \cite{Goodfellow-et-al-2016-dropout}. This loss is defined as 
\begin{equation}
H(p,q) = -\sum_{x}(p(x)\log{(q(x))})
\vspace*{-1mm}
\end{equation}
where $p(x)$ is a probability distribution of true labels and $q(x)$ is a probability distribution of predicted labels. Categorical entropy is a good default choice and it has an advantage that different classes can be easily preferred by giving different weights for the classes.

We used the Adam optimiser \cite{kingma2014adam} for training the model to avoid challenging points in the optimisation space.

All hyper parameters are shown in the table \ref{table:hyperparameters}.

\setlength\e{\dimexpr .20\columnwidth -2\tabcolsep}
\setlength\ee{\dimexpr .55\columnwidth -2\tabcolsep}
\begin{table}[h]
    \centering
    \caption{Hyperparameters of the MLP classifier}
    \label{table:hyperparameters}
    \begin{tabular}[width=\columnwidth]{ p{\ee} p{\e} } 
    \textbf{Parameter} & \textbf{Value } \\
    \hline
    Batch size & 256 \\ 
    \hline
    Epoch count & 1000 \\ 
    \hline
    Dropout probability & 10 \% \\ 
    \hline
    $\alpha$ (learning rate) & 0.001 \\
    \hline
    $\beta_1$ (exp decay for momentum) & 0.9 \\
    \hline
    $\beta_2$ (exp decay for momentum) & 0.999 \\
    \hline
    $\epsilon$ (stability constant) & $10^{-8}$ \\
    \hline
    Initial decay & no decay \\
    \end{tabular}
\end{table}

\section{Experiments}\label{sec:Experiments}

We divided the data set into training and validation set with share of 75 \% and 25 \% respectively. The RFC worked relatively well without requiring any tuning of hyper parameters. The classification accuracy obtained for the training set was around 98 \% and for the validation set up to 88 \%. Thus, 
the RFC tends to slightly over-fit the training data. 

\begin{figure}[t]
\centering
\includegraphics[width=0.9\linewidth]{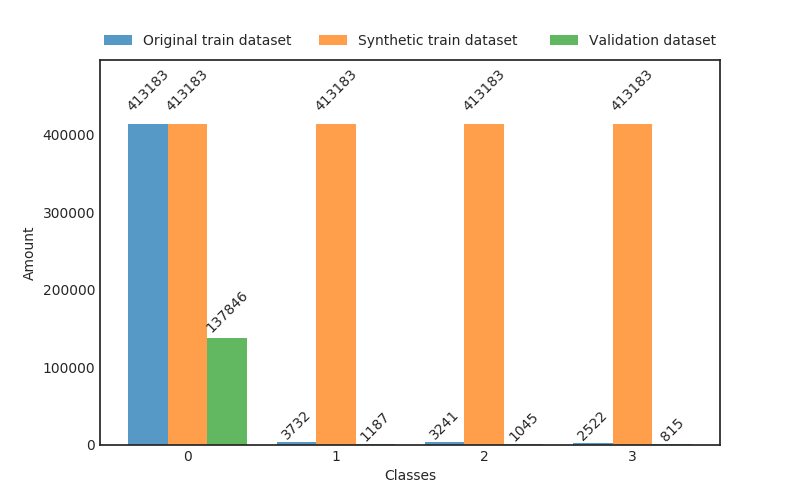} 
\caption{Histogram of classes in original dataset, after generating new samples synthetically and in validation dataset. The original dataset is very imbalanced but after generating synthetic samples by SMOTE, training dataset is completely balanced.}
\label{fig:y_hist_syntetic}
\vspace{-10pt}
\end{figure}

The MLP network required significantly more efforts for tuning the hyperparameters (see Table \ref{table:hyperparameters}). Moreover, and in order to balance the training data, we combined the MLP classifiers with the synthetic minority over-sampling technique (SMOTE) \cite{chawla2002smote}. The method generates new training samples in the vicinity of the original training samples by interpolating their $k=5$ nearest neighbours (in the feature space) as following: 
\begin{equation}
x_{new} = x_i + \lambda \times(x_{zi} - x_i)
\vspace*{-1mm}
\end{equation}
where $x_i$ is an original minority class sample, $x_{zi}$ is one of $x_i$'s $k$ nearest neighbour and $\lambda$ is random variable drawn uniformly from the interval $[0,1]$. The synthetic data set generated by SMOTE contains data points with a balanced distribution of classes (see figure \ref{fig:y_hist_syntetic}).

\begin{figure}[t]
\centering
\includegraphics[width=0.9\linewidth]{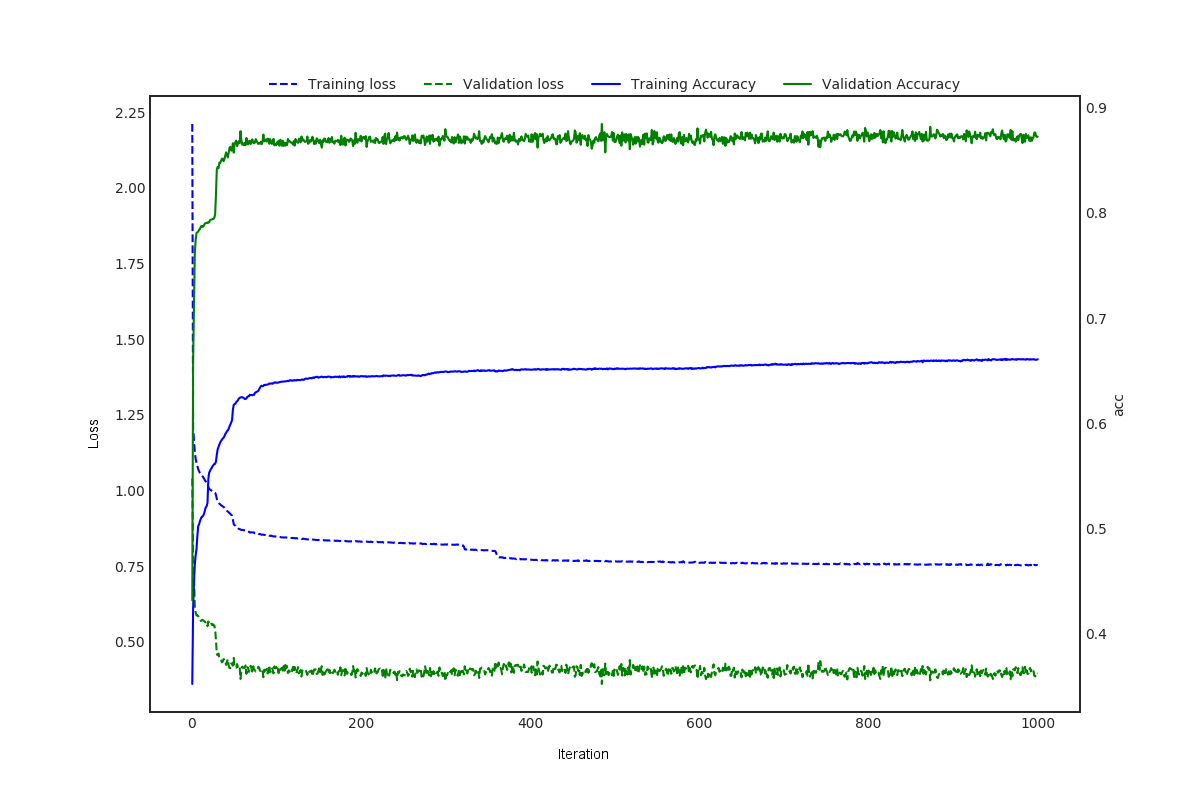} 
\caption{Training and validation metrics while training the MLP network. Validation performance is significantly better than training performance since training dataset contains synthetically generated samples which do not exist in the validation set.}
\label{fig:nn_perf}
\vspace{-10pt}
\end{figure}

Batch size was compromised to provide enough performance in sufficient time. Amount of hidden layers, size of hidden layers and learning rate was searched by trial and error. Training and validation loss and accuracy are plotted in figure \ref{fig:nn_perf}. There is no sign of over-fitting and thus quite low dropout  probability was used for both dropout layers \cite{Goodfellow-et-al-2016-dropout}.

As mentioned in section \ref{sec:Problemsetup}, the data contained large amount of incomplete samples. Intuitive assumption would say, that filtering those samples would be beneficial to gain better results. We created a new dataset $\mathbf{X_{filt}}$ from samples which contained all parameters and used that to train the classification methods. The new dataset contained 563 571 samples. Filtering clean samples from the dataset did not improve the results.




\section{Results}\label{sec:Results}

\begin{figure}[H]
\centering
\includegraphics[width=0.7\linewidth]{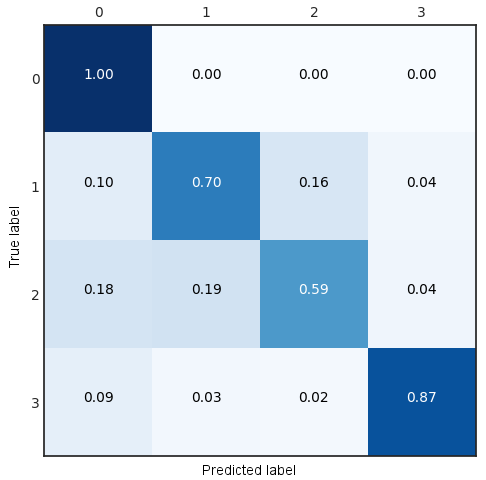}
\caption{Confusion matrix of Random Forest Classifier}
\label{fig:rfc}
\end{figure}

\begin{figure}[ht]
\centering
\includegraphics[width=0.7\linewidth]{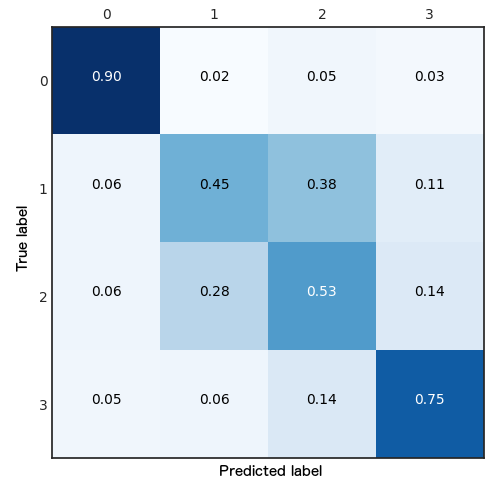}
\caption{Confusion matrix of MLP}
\label{fig:confusion_matrix_normalised}
\end{figure}

The results are shown in table \ref{table:results}. In the end, RFC performed better for predicting amount of damage. It is notable that AUC, loss or accuracy do not catch differences very well as class 0 dominates all metrics. Thus differences between methods can be best shown in confusion matrices (figure \ref{fig:rfc} and \ref{fig:confusion_matrix_normalised}).  

A performance of RFC is excellent for class 0 (no damage) with over 99 \% accuracy and good for class 3 (most harmful) with 87 \% accuracy. The classifier has little problems with distinguishing classes 1 and 2 as it tends to underestimate a damage potential of storms. The results of MLP classifier was significantly worse. While MLP provided 90 \% accuracy for class 0 and 75 \% accuracy for class 3, accuracy for classes 1 and 2 are only near 50 \%.

Classification is fast with both classifiers. For example, 221 506 samples can be classified in about 5 second with two 3,3 GHz i5 CPUs. While optical flow is a lightweight algorithm, DBSCAN method has a $\mathcal{O}(n\log{}n)$ computational cost \cite{ester1996density}, which may be challenging during ``active days''.


\setlength\e{\dimexpr .17\columnwidth -2\tabcolsep}
\setlength\ee{\dimexpr .38\columnwidth -2\tabcolsep}
\begin{table}[h]
    \centering
    \caption{Metrics for different methods. 'MLP SMOTE' means MLP neural network with synthetic samples and 'MLP $\mathbf{X_{filt}}$' means MLP with synthetic samples without samples with missing values. Micro average for F1 score is calculated by counting the total true positives, false negatives and false positives over all classes.}
    \label{table:results}
    \begin{tabular}[width=\columnwidth]{ p{\ee} p{\e} p{\e} p{\e} } 
    \textbf{Metrics} & \textbf{MLP SMOTE} & \textbf{MLP $\mathbf{X_{filt}}$} & \textbf{RFC}\\
    \hline
     AUC & 0.96 & 0.95 & 0.99\\
    \hline
     Validation accuracy & 89 \% & 85 \% & 96 \%\\
    \hline
     F1 score \\ micro average & 71 \% & 85 \% & 99 \%\\
    \end{tabular}
    \vspace{-10pt}
\end{table}

\section{Conclusions}
\label{sec:Conclusion}

This paper studied the application of RFC and MLP classifiers to the problem of predicting power grid outages caused by hazardous storm cells. Some illustrative numerical experiments based on weather data collected by FMI indicated that RFC can outperform deep MLP in predicting the amount of damage caused by storm cells. 

This work suggests several interesting avenues for future research.  One promising direction is to use more advanced models and methods for the training data, e.g., times series models and recurrent neural networks. So far, we also used only very basic methods for coping with missing data (just replace by zero) and imbalanced training data (using SMOTE). It would be interesting to apply more advanced techniques for coping with imbalanced data, e.g., the  ``Rare-Transfer'' algorithm \cite{al2016transfer}.

Currently Random Forest Classifier is used in operational application.

\pagebreak

\bibliographystyle{IEEEbib}
\bibliography{strings,refs}

\end{document}

%% file: param_table.tex
\setlength\e{\dimexpr .35\columnwidth -2\tabcolsep}
\setlength\ee{\dimexpr .65\columnwidth -2\tabcolsep}
\begin{table}[h]
    \centering
    \caption{Used input features}
    \label{table:input_data}
    \begin{tabular}[width=\columnwidth]{ p{\e} p{\ee} } 
    \textbf{Feature} & \textbf{Explanation} \\
    \hline
    Area & Area covered by the storm cell \\ 
    \hline
    Age & Age of the storm \\ 
    \hline
    Lightning density & Lightning density under storm cell \\ 
    \hline
    Max DBZ & Maximum radar reflectivity of the storm cell (spatially). Represents maximum rain intensity. \\
    \hline
    Min DBZ & Minimum radar reflectivity of the storm cell (spatially). Represents minimum rain intensity. \\
    \hline
    Mean DBZ & Mean radar reflectivity of the storm cell (spatially) \\
    \hline
    Median DBZ & Median radar reflectivity of the storm cell (spatially)\\
    \hline
    Std of DBZ & Standard deviation of radar reflectivity of the storm cell (spatially)\\
    \hline
    Lat & Storm center latitude \\
    \hline
    Lon & Storm center longitude \\
    \hline
    Temperature & Air temperature from ground\\ observations \\
    \hline
    Pressure &  Air pressure from ground\\ observations \\
    \hline
    Wind speed &  Wind speed from ground\\ observations \\
    \hline
    Wind direction &  Wind direction from ground\\ observations \\
    \hline
    Precipitation amount &  Precipitation amount from ground observations \\
    \hline
    Snow depth &  Snow depth from ground\\ observations \\
    \end{tabular}
\end{table}

%% file: nn_struct.tex
\def\layersep{2.5cm}
\def\relu{\begin{tikzpicture}[trim axis left, trim axis right, baseline]
          \begin{axis}[axis lines=none,ylabel shift=-0.1cm]
          \addplot+[mark=none,red,domain=-1:0] {0};
          \addplot+[mark=none,red,domain=0:2] {x};
          \end{axis}
          \end{tikzpicture}
          }
\newdimen\XCoord
\newdimen\YCoord
\newcommand*{\ExtractCoordinate}[1]{\path (#1); \pgfgetlastxy{\XCoord}{\YCoord};}%
\begin{tikzpicture}[shorten >=1pt,->,draw=black!50, node distance=\layersep, baseline=(current bounding box.center)]
    \tikzstyle{every pin edge}=[<-,shorten <=1pt]
    \tikzstyle{neuron}=[circle,fill=black!25,minimum size=17pt,inner sep=0pt]
    \tikzstyle{input neuron}=[neuron, fill=green!50];
    \tikzstyle{output neuron}=[neuron, fill=red!50];
    \tikzstyle{hidden neuron}=[neuron, fill=blue!50];
    \tikzstyle{activation neuron}=[neuron, fill=purple!50];
    \tikzstyle{annot} = [text width=8em, text centered]
    \tikzstyle{probState}=[draw=blue!50,
                    fill=white,
                    scale=0.1,
                    minimum size=5cm,
                    minimum width=7cm]
    
    \newcounter{count}
    \setcounter{count}{1}
    \foreach \name / \y in {1,...,3}
        \stepcounter{count}
        \node[input neuron, pin=left:Input] (I-\name) at (0,-\value{count}) {};
    
    \stepcounter{count}    
    \node at (0,-\value{count}) {\ldots};
    
    \foreach \name / \y in {4,...,6}
        \stepcounter{count}
        \node[input neuron, pin=left:Input] (I-\name) at (0,-\value{count}) {};
    
    \def\l{2.5cm}
    \setcounter{count}{0}
    \foreach \name / \y in {1,...,4}
        \stepcounter{count}
        \node[hidden neuron] (H1-\name) at (\layersep,-\value{count}) {};
    
    \stepcounter{count}    
    \node at (\l,-\value{count}) {\ldots};
    
    \foreach \name / \y in {5,...,8}
        \stepcounter{count}
        \node[hidden neuron] (H1-\name) at (\layersep,-\value{count}) {};
        
    \stepcounter{count}
    \node[probState] (A1) at (\l, -\value{count}) {\relu};
    
    \def\l{5cm}
    \setcounter{count}{1}
    \foreach \name / \y in {1,...,3}
        \stepcounter{count}
        \node[hidden neuron] (H2-\name) at (\l,-\value{count}) {};
    
    \stepcounter{count}    
    \node at (\l,-\value{count}) {\ldots};
    
    \foreach \name / \y in {4,...,6}
        \stepcounter{count}
        \node[hidden neuron] (H2-\name) at (\l,-\value{count}) {};
        
    \stepcounter{count}
    \node[probState] (A2) at (\l, -\value{count}) {\relu};

    \coordinate [] (D1) at ($(H1-1)!0.5!(H2-1)$);
    \def\y{-9}
    \ExtractCoordinate{D1}
    \coordinate [] (D1A) at (\XCoord, \y);
    \draw [-, dashed] (\XCoord, -1) -- (\XCoord, \y);
    
    \def\l{7.5cm}
    \setcounter{count}{2}
    \foreach \name / \y in {1,...,2}
        \stepcounter{count}
        \node[hidden neuron] (H3-\name) at (\l,-\value{count}) {};
    
    \stepcounter{count}    
    \node at (\l,-\value{count}) {\ldots};
    
    \foreach \name / \y in {3,...,4}
        \stepcounter{count}
        \node[hidden neuron] (H3-\name) at (\l,-\value{count}) {};
    
    \stepcounter{count}
    \node[probState] (A3) at (\l, -\value{count}) {\relu};
    
    \coordinate [] (D2) at ($(H2-1)!0.5!(H3-1)$);
    \def\y{-8}
    \ExtractCoordinate{D2}
    \coordinate [] (D2A) at (\XCoord, \y);
    \draw [-, dashed] (\XCoord, -2) -- (\XCoord, \y);
    
    \def\l{10cm}
    \setcounter{count}{2}
    \foreach \name / \y in {1,...,4}
        \stepcounter{count}
        \path[yshift=-0.5cm]
            node[hidden neuron] (H4-\name) at (\l,-\value{count}) {};
    
    \stepcounter{count}
    \path[yshift=-0.5cm]
        node[probState] (A4) at (\l, -\value{count}) {};

    \def\l{12.5cm}
    \node[output neuron,pin={[pin edge={->}]right:Output}] (O) at (\l, -5cm) {};

    \foreach \source in {1,...,6}
        \foreach \dest in {1,...,8}
            \path (I-\source) edge (H1-\dest);
    
    \foreach \source in {1,...,8}
        \foreach \dest in {1,...,6}
            \path (H1-\source) edge (H2-\dest);    
            
     \foreach \source in {1,...,6}
        \foreach \dest in {1,...,4}
            \path (H2-\source) edge (H3-\dest);    

     \foreach \source in {1,...,4}
        \foreach \dest in {1,...,4}
            \path (H3-\source) edge (H4-\dest);  

    \foreach \source in {1,...,4}
        \path (H4-\source) edge (O);

    \node[annot,above of=I-1, node distance=1cm] {Input [16x1]};
    
    \node[annot,above of=H1-1, node distance=1cm] (hl) {Fully\ connected [20x1]};
    \node[annot,above of=H2-1, node distance=1cm] (hl) {Fully\ connected [16x1]};
    \node[annot,above of=H3-1, node distance=1cm] (hl) {Fully\ connected [8x1]};
    \node[annot,above of=H4-1, node distance=1cm] (hl) {Fully\ connected [4x1]};
    
    \node[annot,below of=A1, node distance=0.5cm] (hl) {ReLU};
    \node[annot,below of=A2, node distance=0.5cm] (hl) {ReLU};
    \node[annot,below of=A3, node distance=0.5cm] (hl) {ReLU};
    \node[annot,below of=A4, node distance=0.5cm] (hl) {SoftMax};
    
    \node[annot, below of=D1A, node distance=0.5cm] (hl) {Dropout\\ 10 \%};
    \node[annot, below of=D2A, node distance=0.5cm] (hl) {Dropout\\ 10 \%};
    
\end{tikzpicture}